\def\year{2020}\relax
\documentclass[letterpaper]{article} 
\usepackage{aaai20}  
\usepackage{times}  
\usepackage{helvet} 
\usepackage{courier}  
\usepackage[hyphens]{url}  
\usepackage{graphicx} 
\newcommand{\citet}[1]{\citeauthor{#1} \shortcite{#1}}

\usepackage{array}
\usepackage{multirow}
\newcolumntype{C}[1]{>{\centering\let\newline\\\arraybackslash\hspace{0pt}}p{#1}}

\usepackage{amsmath}
\usepackage{verbatim}

\newcommand{\ourdslong}{IBM-ArgQ-Rank-30kArgs}
\newcommand{\ourds}{IBM-Rank-30k}

\title{A Large-scale Dataset for Argument Quality Ranking: Construction and Analysis}
\author {
 Shai Gretz\thanks{\ \ These authors equally contributed to this work.}, Roni Friedman$^{*}$, Edo Cohen-Karlik$^{*}$, Assaf Toledo$^{*}$, Dan Lahav,\\
\Large \textbf{Ranit Aharonov and Noam Slonim}\\
 IBM Research
}
\author{
Shai Gretz\thanks{\ \ These authors equally contributed to this work.}, Roni Friedman$^{*}$, Edo Cohen-Karlik$^{*}$, Assaf Toledo$^{*}$, Dan Lahav,\\
\Large \textbf{Ranit Aharonov and Noam Slonim}\\
IBM Research\\
\{avishaig,roni.friedman-melamed,noams\}@il.ibm.com\\
\{edo.cohen,assaf.toledo,dan.lahav,ranit.aharonov\}@ibm.com
}

\begin{document}
\maketitle

\begin{abstract}

Identifying the quality of free-text arguments has become an important task in the rapidly expanding field of computational argumentation. In this work, we explore the challenging task of argument quality ranking. To this end, we created a corpus of $30{,}497$ arguments carefully annotated for point-wise quality, released as part of this work. To the best of our knowledge, this is the largest dataset annotated for point-wise argument quality, larger by a factor of five than previously released datasets. Moreover, we address the core issue of inducing a labeled score from crowd annotations by performing a comprehensive evaluation of different approaches to this problem. In addition, we analyze the quality dimensions that characterize this dataset. Finally, we present a neural method for argument quality ranking, which outperforms several baselines on our own dataset, as well as previous methods published for another dataset.

\end{abstract}

\section{Introduction}
\label{intro}

Computational Argumentation is a rapidly emerging discipline within the Natural Language Processing community \cite{ACL-16-tutorial}, dealing with various sub-tasks such as argument detection \cite{LippiTorroni,ein-dor19}, stance detection \cite{bar-haim17} and argument clustering \cite{GurevychClustering19}. 

Recently, IBM introduced \textit{Project Debater}, the first AI system able to debate humans on complex topics. The system participated in a live debate against a world champion debater, and was able to mine arguments, use them for composing a speech supporting its side of the debate, and also rebut its human competitor.\footnote{For more details: \url{https://www.research.ibm.com/artificial-intelligence/project-debater/live/}} The underlying technology is intended to enhance decision-making.

More recently, IBM also introduced \textit{Speech by Crowd}, a service which supports the collection of free-text arguments from large audiences on debatable topics to generate meaningful narratives \cite{Toledo19}. An important sub-task of this service is automatic assessment of argument quality, which is the focus of the present work. Detecting argument quality is a prominent task due to its importance in automated decision making \cite{BenchCapon09}, argument search \cite{WachsmuthWeb}, and writing support \cite{stabGurevych14}.

Earlier work on assessing argument quality relied on a comparative {\it pair-wise} approach, aiming to identify the higher-quality argument within each pair of arguments \cite{Gurevych16,Gurevych18}. 
Recently \citet{Toledo19} 
proposed a {\it point-wise\/} argument quality prediction scheme, which scales linearly with data size. Correspondingly, here we focus on 
this paradigm 
which is clearly less demanding, especially in scenarios where many arguments should be considered.

A major contribution of this work is introducing a novel dataset of arguments, carefully annotated for point-wise quality, \textit{\ourdslong{}}, referred henceforth as \textit{\ourds{}}. The dataset includes around $30k$ arguments, $5$ times larger than the largest annotated point-wise data released to date \cite{Toledo19}. Similarly to \citet{Toledo19}, arguments were collected actively -- as opposed to being extracted from debate portals \cite{Gurevych16} -- with strict length limitations, accompanied by extensive quality control measures.

Although a continuous argument quality score seems natural, asking annotators to provide a continuous score per argument will probably introduce a subjective scale that varies from one annotator to another, 
hindering downstream analysis. Instead, we take 
a simplified approach, asking each annotator to answer a binary question per argument, indicating if 
its quality 
is satisfactory in a particular context. The question remains, of how to extract a continuous quality score out of the binary annotations provided by many annotators. \citet{Toledo19} took a straightforward simple-average approach; here, we provide an extensive comparison between potential scoring functions, analyzing the differences between these models, and their impact on training and evaluating a learning algorithm. 

While an exact definition of argument quality is potentially elusive, it seems clear that it is a function of various linguistic phenomena. As exemplified in Table \ref{table:examplesSBC}, low quality can be manifested by dimensions such as bad grammar and low clarity (Row $3$), or lack of impact and relevance (Row $4$). 
In contrast, 
high quality arguments are typically clear, relevant, and with 
high impact (Rows $1$ and $2$). A model that aims to automatically infer argument quality should take such subtleties into account. A recent development 
in this context is that of deep contextual language models, such as ELMo \cite{Peters:2018} and BERT \cite{BERT18}. Due to its bidirectional nature, BERT achieves remarkable results when fine-tuned to different tasks without the need for specific modifications per task. As part of this work, we introduce various neural methods that exploit the value of BERT for our task. 
In particular, we suggest 
a model that outperforms several baselines on our data. Our experimental results further indicate that this method is either comparable to or outperforms state-of-the-art methods on previously released data.

\begin{table}[ht]
\begin{center}
\scriptsize
\begin{tabular}{ p{0.35\linewidth}|c|c  }
 \textbf{Argument} & \textbf{Topic} & \textbf{Label}\\
 \hline
 the interest rates are too high and trap people in debt & Payday loans should be banned & $1$ \\
 \hline
  racial profiling unfairly targets minorities and the poor & We should end racial profiling & $1$\\
 \hline
 we should subsidize student loans for reach excelent education & We should subsidize student loans & $0.05$\\
 \hline
 i think the same as you, they should ban & Payday loans should be banned & $0.09$ \\
 \end{tabular}
 \end{center}
 \caption{Examples of high (rows 1-2) and low (rows 3-4) quality arguments from the \textit{\ourds{}} dataset. The label is a weighted aggregation of annotations.}
\label{table:examplesSBC}
\end{table}

The main contributions of our work are: (1) Introducing a carefully annotated argument quality dataset which is the largest of its kind; (2) Conducting extensive analysis of different approaches to induce a quality label from given binary annotations; (3) Proposing a BERT-based method to predict argument quality, and report the results of extensive experiments that convey the potential of this method. 

\section{Related Work}

Assessing argument quality is a long standing challenge. For centuries there has been a multi-disciplinary effort to define and research aspects of quality in argumentation \cite{Aristotle07,Walton2008,perelman69new}. A core issue in the field is the presumed subjectivity of the task at hand. There have been several practical and theoretical approaches on how to overcome the supposed lack of objectivity.

\citet{swanson2015argument} approach argument quality as a point-wise ranking task, with the goal of selecting argument segments that clearly express an argument facet in a given dialogue. Arguments are labeled by a real value in the range of $[0,1]$, where a score of $1$ indicates that an argument can be easily interpreted. They then develop an automatic regression method using these labels. Their corpus, which we refer to henceforth as \textit{SwanRank}, contains $5.3k$ labeled arguments.

An alternative approach to assess arguments is to focus on their relative \textit{convincingness}, by comparing pairs of arguments with similar stance. This approach is introduced in \citet{Gurevych16,Gurevych18}, and further assessed in \citet{PotashPairs2017}, \citet{gleizeACL2019}, and \citet{potashRanking2019}. As part of their work, \citet{Gurevych16} introduce two datasets: \textit{UKPConvArgRank} (henceforth, \textit{UKPRank}) and  \textit{UKPConvArgAll}, which contain $1k$ and $16k$ arguments and argument-pairs, respectively. In their work, the point-wise scores are induced from the pair-wise labels, rather than annotated directly. \citet{gleizeACL2019} focus of ranking convincingness of \textit{evidence}. Their solution is based on a Siamese neural network, which outperforms the results achieved in \citet{Gurevych18} on the \textit{UKP} datasets, as well as several baselines on their own dataset, \textit{IBM-ConvEnv}. \citet{PotashPairs2017} present a method that is based on representing an argument by the sum of its token embeddings, extended in \citet{potashRanking2019} to include a Feed Forward Neural Network. These two works outperform \citet{Gurevych18} on the \textit{UKP} datasets for both the pair-wise and point-wise tasks, respectively.

\citet{durmus2019} present a new dataset comprised of over $47k$ claims in $471$ topics from the website \textit{kialo.com}, aimed at evaluating the effect of pragmatic and discourse context when determining argument quality. They propose models to predict the \textit{impact} value of each claim, as determined by the users of the website. Their dataset is somewhat different from ours as it focuses on argument impact, rather than overall quality, and doing so in the context of an argumentative structure, instead of independently. In addition, their impact values are based on spontaneous input from users of the website, whereas our dataset was carefully annotated with clear guidelines. Still, it further highlights the importance of this field.\footnote{The work by \citet{durmus2019} was published after our submission, therefore we were not able to fully address it.}

\citet{Toledo19} consider both point-wise and pair-wise quality approaches, as well as the interaction between them. They introduce two datasets: \textit{IBMRank}, which contains $5.3k$ point-wise labeled arguments ($6.3k$ before cleansing) and \textit{IBMPairs}, which contains $9.1k$ labeled argument-pairs ($14k$ before cleansing). Arguments in their datasets, collected in the context of Speech by Crowd experiments, are suited to the use-case of civic engagement platforms, giving premium to the usability of an argument in oral communication. Our dataset differs in three respects: (1) our dataset is larger by a factor of $5$ compared to previous datasets annotated for point-wise quality; (2) our data were collected mainly from crowd contributors that presumably better represent the general population compared to targeted audiences such as debate clubs; (3) we performed an extensive analysis of argument scoring methods and introduce superior scoring methods that consider annotators credibility without removing them entirely from the labeled data, as is done in \citet{Toledo19}. 

\section{\ourds{} Dataset}

In the next two sections, we present the creation of the \textit{\ourds{}} dataset. First, we describe the process of argument collection. We then move to describing how 
arguments were annotated for quality. 
Finally, we discuss and analyze how we derive point-wise continuous quality labels from binary annotations, by conducting a comprehensive comparison between different scoring functions. We release this dataset as part of this work.\footnote{\url{http://ibm.biz/debater-datasets}}

\subsection{Argument Collection}
\label{argCollection}
For the purpose of collecting arguments for the \textit{\ourds{}} dataset, we conducted a crowd annotation task, using the Figure Eight platform.\footnote{http://figure-eight.com/} A small portion of arguments ($8.6\%$) was also collected from expert annotators who work closely with our team. 
We selected $71$ common controversial topics, for which arguments were collected (e.g., \textit{We should abolish capital punishment}).

We follow similar guidelines to that presented in \citet{Toledo19}. Annotators were presented with a single topic each time, and asked to contribute one supporting and one contesting argument for it, requiring arguments to be written using original language. To motivate high-quality contributions, contributors were informed they will receive extra payment for high quality arguments, as determined by the subsequent argument quality labeling task (Section \ref{argQualityLabeling}). It was explained that an argument will be considered as a high-quality one, if a person preparing a speech on the topic will be likely to use this argument \emph{as is} in her speech.

Similarly to \citet{Toledo19}, we place a limit on argument length - a minimum of $35$ characters and a maximum of $210$ characters. In total, we collected $30{,}497$ arguments from $280$ contributors, each contributing no more than $6$ arguments per topic.

\subsection{Argument Quality Annotations Collection}
\label{argQualityLabeling}
In this section we describe the argument quality annotation process, performed for all collected arguments. As above, we used the Figure Eight platform, with $10$ annotators 
per argument. 
Following \citet{Toledo19}, annotators were presented with a binary question per argument, asking if they would recommend a friend to use that argument \emph{as is} in a speech supporting/contesting the topic, regardless of personal opinion. In addition, annotators were asked to mark the stance of the argument towards the topic (\textit{pro} or \textit{con}).

To monitor and ensure the quality of the collected annotations, we employed the following measures introduced in \citet{Toledo19}:

\textbf{Test Questions.} Before the labeling of $1/5$ of the arguments,
a hidden test question about the stance of the argument towards the topic was presented, aimed to verify the annotator is reading the argument carefully. Annotators that failed more than $20\%$ of the test questions were removed from the task, and their judgments were ignored. Typically $5\%-10\%$ of the contributors were removed from each sub-task due to this reason. 

\textbf{Annotator-reliability score}. Defined in \citet{Toledo19} as the Annotator-$\kappa$ for a single score (and task-average-$\kappa$ averaged on all valid scores) and denoted here as $Annotator$-$Rel$. This score was used both to monitor tasks in real time, and as a basis for the weighted score function described in Section \ref{qualityScore}. It is obtained by averaging all pair-wise $\kappa$ for a given annotator, with other annotators that share at least $50$ common judgements. Annotators who do not share at least $50$ common judgments with at least $5$ other annotators, do not receive a value for this score.

The average of valid annotators reliability scores on the quality annotations is $0.12$. This task reproduces the task from \citet{Toledo19}, and as established there, such an average is acceptable due to the subjectivity of the task.\footnote{In \citet{Toledo19} it was $0.1$.} Among other things, \citet{Toledo19} rely on the task-average-$\kappa$ of the stance annotations in the same task, which was $0.69$ in that work, and here it is $0.83$.

To further ensure high quality annotations, rather than introducing the task to any crowd worker, we used a selected group of $600$ crowd annotators, which had high Annotator-reliability in past tasks of our team.

\section{Deriving an Argument Quality Score from Binary Annotations} \label{qualityScore}
As mentioned in section \ref{argQualityLabeling}, for each argument several annotators answered a binary question regarding its quality. We chose this format to simplify the annotation process of this 
subjective question, 
aiming to avoid 
an additional subjective element - the scale. However, we still need 
to provide a single continuous score per 
argument, reflecting 
its quality, 
that can be compared with the scores of other arguments. 
To that end, we evaluate approaches
to derive such a quality score, on a bounded scale, from a set of binary annotations.

\subsection{Quality Scoring Functions}

First, we describe two scoring functions. Each function provides the likelihood of the positive label, between $0$ (bad quality) and $1$ (good quality).

\textbf{MACE probability (\textit{MACE-P})} - \citet{Gurevych16} suggested MACE \cite{hovyMACE} as a scoring function for the quality of an argument based on crowd annotations. MACE is an unsupervised item-response generative model which predicts the probability for each label given the annotations.
MACE also estimates a reliability score for each annotator which it then uses to weigh this annotator's judgments. We use the probability MACE outputs for the positive label as the \textit{MACE-P} scoring function.

\textbf{Weighted-Average (\textit{WA})} - We suggest to use a weighted-average score to incorporate annotator-reliability, in the spirit of MACE.
This is designed to decrease the influence of non-reliable annotators on the final quality score, thus providing an intuitive and gradual form of data cleansing.
For each argument $a$, we define $P_a$ as the set of annotators who labeled it as positive, and $N_a$ as the set of annotators who labeled it as negative.
The \textit{WA} score of an argument $a$ is defined as:
\[
Score(a) = \frac{\sum_{Annotator_i\in P_a}^{}Annotator\text{-}Rel_i}{\sum_{Annotator_j\in N_a+P_a}^{}Annotator\text{-}Rel_j}
\]

\begin{figure}[htb]
\begin{center}
\includegraphics[width=.95\columnwidth]{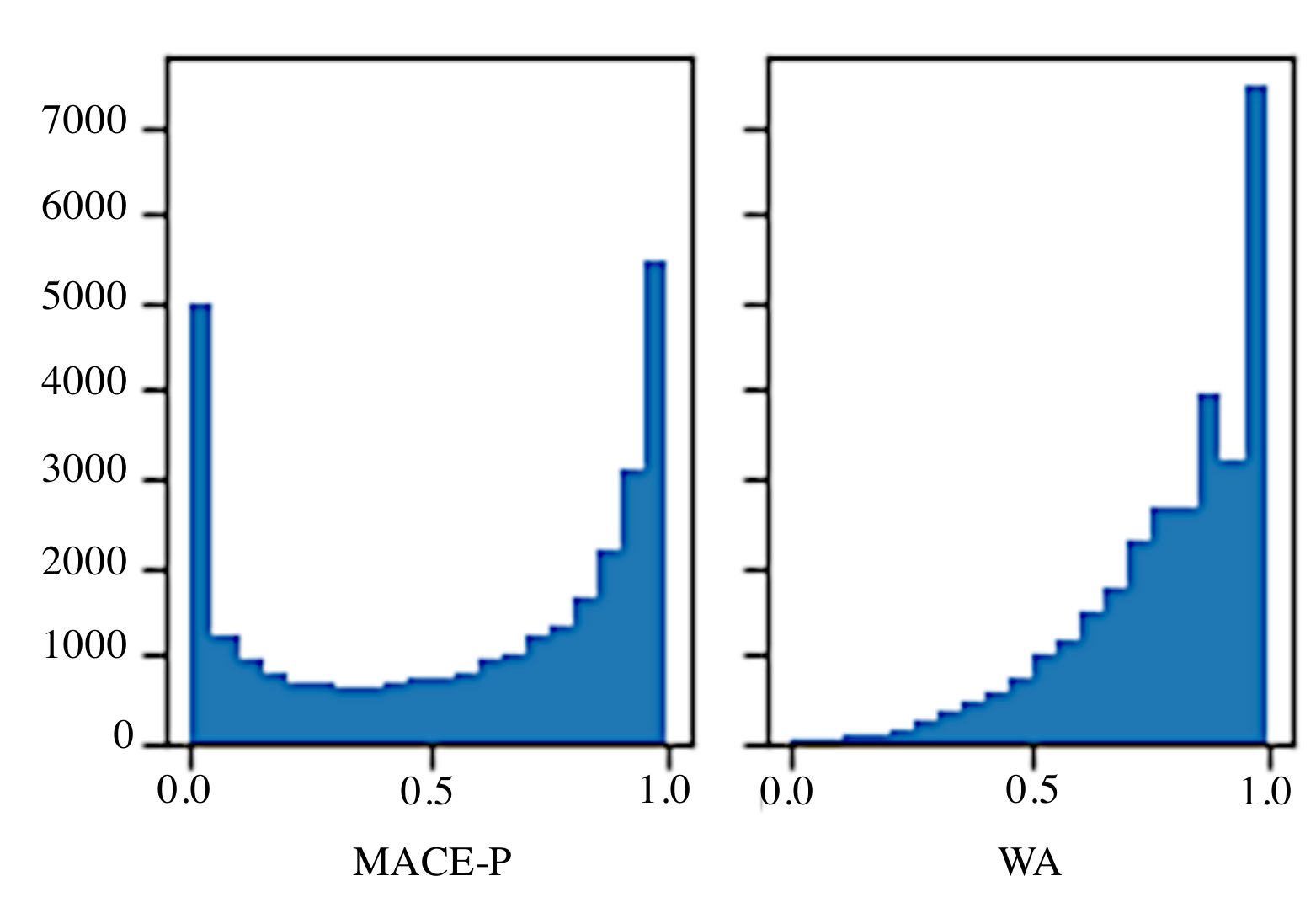}
\caption{Histogram of arguments according to \textit{MACE-P} and \textit{WA} quality scores. X axis - quality scores. Y axis - counts of arguments in \textit{\ourds{}}.}
\label{fig:histWA_MACE}
\end{center}
\end{figure}

There is a clear distinction between the distribution of scores obtained by \textit{WA} and \textit{MACE-P} scoring functions (see Figure \ref{fig:histWA_MACE}).
\textit{WA} outputs values close to $0$ or $1$ only if there is a strong annotation consensus. As generally there are more positive annotations in our data, the histogram of this scoring function is skewed towards $1$, with an almost linear decrease. On the other hand, MACE assigns probabilities to both labels. As a result, the quality scores lean strongly to both extreme values, creating a U-shaped histogram.

\subsection{Comparing Scoring Functions}
Next, we compare 
the quality scoring functions described above
via 
the following three experiments.

\subsubsection{Disagreements in choosing the better argument} 
We first ask which of the two methods is `correct' 
more often when examining the cases in which given a pair of arguments, they disagree on which argument should get a higher score. We created all possible argument pairs from \textit{\ourds{}}, such that arguments in the same pair are taken from the same topic and have the same stance towards it (to avoid annotator bias).\footnote{Arguments were selected from the training set, which is detailed in Section \ref{experiments}.} 

We focus on the set of pairs in which the two methods disagree on the preferred argument, i.e. the one that received a higher score (about $20\%$ of the pairs). 
We sample $850$ such disagreement pairs and send them for \textit{pair-wise} annotation, i.e., asking the annotators to choose the better of the two arguments. Each pair was annotated by $7$ annotators from our selected group of crowd annotators. We discarded pairs for which the agreement between annotators was less than $70\%$ (this filtered out $27\%$ of the pairs). 
In $55\%$ of the remaining 
pairs the annotators chose the argument preferred by \textit{MACE-P}, making this method somewhat more correlated with the pair-wise judgments. Both methods were also compared to simple-average, the scoring method applied in \citet{Toledo19}. In $61\%$ of the pairs that differ between simple-average and \textit{MACE-P}, annotators chose the argument preferred by the latter. In $59\%$ of the pairs that differ between simple-average and \textit{WA}, annotators chose the argument preferred by \textit{WA}. As the simple-average method was found inferior, it was omitted from subsequent evaluation experiments.

\subsubsection{Agreement with pair-wise annotation} 
In this experiment, we present each scoring function with pairs of arguments, and compare their preferred arguments with a gold standard obtained by pair-wise annotation, following the consistency evaluation performed in \citet{Toledo19}.
Similar to the previous experiment, we generate all argument pairs (of the same topic and stance). Next, we bin all pairs into four sets by the size of the delta between the scores of their arguments (e.g., all pairs with score difference between $0.25$ and $0.5$).
From each such \textit{delta bin} we randomly sample $150$ pairs and send them for pair-wise annotation as described in the previous experiment.
We repeat this process for both scoring functions.

We calculate the precision of each scoring function in each delta bin, based on its agreement with the pair-wise annotation.
Table \ref{table:macePairsConsistency} depicts the results for \textit{MACE-P}, and Table \ref{table:WLAPairsConsistency} for \textit{WA}.
The results show that as the score difference increases (larger \textit{delta bin}), precision also increases (for both scoring functions). Interestingly, this tendency is more prominent for \textit{WA}, reaching a perfect match when the difference in point-wise quality scores is higher than $0.75$.\footnote{However, we note that while for \textit{MACE-P} the $>.75$ delta bin holds 
for 
$19.5\%$ of the pairs, for \textit{WA} it holds less than $1\%$.} 
The tables also depict the percent of pairs filtered due to low 
(less than $70\%$) 
agreement between annotators. Interestingly, in the $3$ bins with more than $0.25$ difference in point-wise scores, less pairs are filtered when using \textit{WA}, suggesting that when \textit{WA} is confident about which argument is better, annotators also tend to agree more often, which is not the case for \textit{MACE-P}. 

\begin{table}[htb]
\begin{center}
\begin{tabular}{ccc}
 Delta bin & Filtered pairs & Precision\\
 \hline
 $\leq 0.25$ & $15\%$ & $0.64$\\
 $(0.25, 0.50]$ & $23\%$ & $0.77$\\
$(0.50, 0.75]$ & $14\%$ & $0.81$\\
$> 0.75$ & $13\%$ & $0.88$\\
\hline
 \end{tabular}
 \end{center}
 \caption{Comparing \textit{MACE-P} scoring function preference to gold standard of pair-wise annotation.}
\label{table:macePairsConsistency}
\end{table}

 \begin{table}[htb]
\begin{center}
\begin{tabular}{ccc}
 Delta bin & Filtered pairs & Precision\\
 \hline
 $\leq 0.25$& $15\%$ & $0.66$\\
 $(0.25, 0.50]$& $13\%$ & $0.74$\\
$(0.50, 0.75]$& $6\%$ & $0.90$\\
$> 0.75$& $3\%$ & $1.00$\\
\hline
 \end{tabular}
 \end{center}
 \caption{Comparing \textit{WA} scoring function preference to gold standard of pair-wise annotation.}
\label{table:WLAPairsConsistency}
\end{table}

\subsubsection{Split annotations consistency} A desirable property 
for a scoring function is that it will be relatively consistent with respect to different sets 
of annotators; namely, if we split the binary annotations into two sets and construct
the continuous annotation from each set independently, we will end up with approximately the same score.
 
To examine that, we randomly split the $10$ binary annotations we have for each argument into two sets (each containing $5$ annotations), similar to \citet{habernal-etal-2018-argument}.
We then calculate two quality scores for each argument, 
based on these two 
smaller annotation sets, respectively. 
While \citet{habernal-etal-2018-argument} utilized agreement to measure consistency, we can measure correlation between the two continuous scores.

We note that since each score was calculated on a set which is half the size of the original annotation set, we have less information for assessing the reliability score of each annotator. This fact can harm the accuracy of the scoring functions, as they both utilize annotators reliability scores.
Nevertheless, for both scoring functions 
we considered, 
we find a good correlation between the scores calculated over the two smaller annotation sets. \textit{WA} achieves $0.42$ and $0.36$ Pearson and Spearman correlations, respectively.  \textit{MACE-P} achieves $0.42$ in both. 

In summary, we point out a key difference between the two scoring functions: the tendency of \textit{WA} to present a gradual continuous scale, as opposed to \textit{MACE-P}, which aims at discovering the `true', hence the binary, labels. 
For this reason we tend to prefer \textit{WA} as a scoring function for this task, which is inherently deriving a non-binary score. However, as our experiments do not show a clear preferred function, we utilize both for the evaluation of neural methods in Section \ref{experiments}. For brevity, in the analyses in sections \ref{dimension} and \ref{analysisModels} we only use \textit{WA} scores. Finally, in the dataset we release as part of this work, we include the quality scores of both scoring functions. We conclude that the induction of a quality score on top of existing annotations is not necessarily trivial, and should be carefully considered, as it impacts the scores distribution and the performance of learning algorithms trained on the score.

\section{Analysis of Quality Dimensions}
\label{dimension}

We seek to further explore the consistency and accuracy of the \textit{\ourds{}} dataset. To that end, we employ a quality dimensions model, i.e. a model that decomposes a holistic quality score to several dimensions. Such a model was suggested by \citet{Wachsmuth2017} - a meta-model that was created via a broad literature survey on previous quality models. They decompose quality to $15$ sub-dimensions that aim to determine fine-grained properties of argument quality.

We conducted an annotation task similar in spirit to \citet{Wachsmuth2017}, in the context of the \textit{\ourds{}} dataset. Our goal is to quantitatively assess the reasons for arguments having low or high quality, through the prism of this theoretical model. First, we decided to exclude 5 dimensions from this task, as they lack high potential to embody relevant characteristics over our data.\footnote{Those 5 dimensions are: Appropriateness, Arrangement, Credibility, Emotional Appeal and Reasonableness.} We split the \textit{\ourds{}} dataset to $5$ equally populated bins according to the \textit{WA} quality scores ($1$-$5$, where $1$ is the lowest quality bin), and randomly sample 100 arguments with a uniform distribution over the bins. Each argument was labeled by 3 expert annotators that have extensive background in related tasks of our team. The annotators 
were not aware of the original quality bins. 
We have asked the annotators to annotate each argument according to each of the $10$ dimensions on a scale of $1$-$3$, to stay consistent with the scale offered by \citet{Wachsmuth2017}, and calculate the average for each dimension.

We observe that even though the task is complex, across all dimensions, the arguments from the highest bin achieve a higher average over the middle bins, and arguments from the middle bins achieve a higher score over the lower bins.
The $2$ dimensions that present the largest difference between bins $5$ and $1$ are \textit{Global Relevance} and \textit{Effectiveness}, with an average difference of $0.72$ and $0.64$, respectively. \textit{Global Relevance} asks whether an argument provides information that helps to arrive at an ultimate conclusion regarding the discussed issue, while \textit{Effectiveness} asks whether the argument is effective in helping to persuade in the author's stance. These results suggest that the differences between low and high arguments in the \textit{\ourds{}} dataset are best explained by how related the annotator found the argument to the topic, and how effectively the argument was presented.
Such results corroborate several notions. Firstly, our quality labeling is broadly consistent with a known quality model.
Moreover, the outcome serves as an initial proof of concept of decomposing quality in a large dataset to gain explainabilty. The detection of quality dimensions can be used in a range of applications, from more exact research on what determines quality, to feedback systems that can recommend users 
more precisely 
what they need to improve to advance their argumentation skills.

\section{Argument Quality Ranking}
\label{argQualityPrediction}

We now move to apply the \textit{\ourds{}} dataset to the task of learning to rank the quality of arguments. For this purpose, we evaluate the following methods, which include several neural methods, as well as some simpler baselines.

\textbf{Arg-Length}. Although we placed a strong limit on argument length, it is possible that there is still a bias towards longer arguments. To inspect this, we evaluate a ranking baseline based on an argument's length in characters.

\textbf{SVR BOW}. We evaluate a Support Vector Regression ranker, implemented by the scikit-learn toolkit,\footnote{https://scikit-learn.org/} with an RBF kernel and bag-of-words features, using the most frequent $1000$ tokens in the training set.

\textbf{Bi-LSTM GloVe}. As a simple neural baseline we implement a Bi-LSTM model with self-attention, following the model used in \citet{levy2018}. The model was trained with a dropout of $0.15$, an LSTM layer of size $128$ and an attention layer of size $100$. For input features we used the $300$ dimensional GloVe embeddings \cite{Pennington14glove}.

We use the following three methods based on BERT:

\textbf{BERT-Vanilla} (henceforth, \textbf{BERT-V}). In this network, for each argument, we concatenate the last 4 layers of the \verb|[CLS]| token obtained from BERT's pre-trained model, resulting in a feature vector of size $4 \times 768 = 3072$. The features are passed through a fully-connected hidden layer of size $100$ with ReLU activation, after which we apply a sigmoid activation layer with a single output. 

\textbf{BERT-Finetune} (henceforth, \textbf{BERT-FT}). This method fine-tunes BERT's pre-trained model. The official code repository of BERT\footnote{\url{https://github.com/google-research/bert}} supports fine-tuning to classification tasks, which is done by applying a linear layer on the \verb|[CLS]| token of the last layer of BERT's model, which is then passed through a soft-max layer. The weights of the preceding layers are initialized with BERT's pre-trained model, and the entire network is then trained on the new data. To adapt the fine-tuning process to a regression task, the following were performed: (1) Changing the label type to represent real values instead of integers; (2) Replacing the softmax layer with a sigmoid function, to support a single output holding values in the range of \verb|[0,1]|; (3) Modifying the loss function to calculate the Mean Squared Error of the logits compared to the labels.\footnote{The implementation of BERT fine-tuning in regression mode is adapted from: \url{https://github.com/google-research/bert/issues/160\#issuecomment-445066341.}}

\textbf{BERT-FT}\textsubscript{TOPIC}. We also evaluate the addition of the topic to the input of BERT-FT. The topic is concatenated to the argument, separated by a \verb|[SEP]| delimiter, and the model is fine-tuned as in BERT-FT.

\subsection{Experiments on \textit{\ourds{}}}
\label{experiments}

For the purpose of evaluating our methods on the \textit{\ourds{}} dataset, we split its $71$ topics to $49$ topics for training, $7$ for tuning hyper-parameters and determining early stopping (dev set) and $15$ for test.

We present results for both \textit{WA} and \textit{MACE-P} scoring functions, aiming to shed some more light on their properties. All models were trained for $5$ epochs over the training data, taking the best checkpoint according to the performance on the dev set, with a batch size of $32$ and a learning rate of $2$e-$5$. We calculate Pearson ($r$) and Spearman ($\rho$) correlations on the entire test set.

For significance testing, we use the Williams test \cite{williams59} which evaluates the significance of a difference in dependent correlations \cite{steiger1980tests}. The Williams test has been successfully used in Machine Translation in the evaluation of MT metrics \cite{Graham14} and quality estimation \cite{Graham15}.

\subsection{Results and Discussion}

The results on the \textit{\ourds{}} dataset are presented in Table \ref{table:resultsSBC}. Using BERT-V improves on the Bi-LSTM GloVe method by $.4$-$.6$ points for Pearson correlation, and $.2$-$.6$ points for Spearman correlation. Fine-tuning BERT improves on BERT-V by $.2$-$.4$ points for both correlation measures. Adding the topic adds a statistically significant improvement of $.1$-$.2$ points ($p\ll0.01$ for both correlation measures and quality score methods). Interestingly, when using \textit{MACE-P} scores, the model is able to achieve higher Spearman correlation for most methods. Also, argument length is not an indicator for quality, as evident by the poor performance of the Arg-Length baseline.

\begin{table}[htb]
\begin{center}
\begin{tabular}{ |l|c|c|c|c|  }
 \hline
 \multirow{2}{*} & \multicolumn{2}{c|}{\textit{WA}} & \multicolumn{2}{c|}{\textit{MACE-P}} \\
 \hline
 & $r$ & $\rho$ & $r$ & $\rho$\\
 \hline
 Arg-Length & $.21$ & $.22$ & $.22$ & $.23$\\
 \hline
 SVR BOW & $.32$ & $.31$ & $.33$ & $.33$ \\
 \hline
 Bi-LSTM GloVe & $.44$ & $.41$ & $.43$ & $.42$\\
 \hline
 BERT-V & $.48$ & $.43$ & $.49$ & $.48$\\
 \hline
BERT-FT & $.51$ & $.47$ & $.52$ & $.5$\\\hline
BERT-FT\textsubscript{TOPIC} & $\textbf{.52}$ & $\textbf{.48}$ & $\textbf{.53}$ & $\textbf{.52}$\\\hline
 \end{tabular}
 \end{center}
 \caption{Correlations on the \textit{\ourds{}} test set.
 }
\label{table:resultsSBC}
\end{table}

\begin{figure}[htb]
\begin{center}
\includegraphics[]{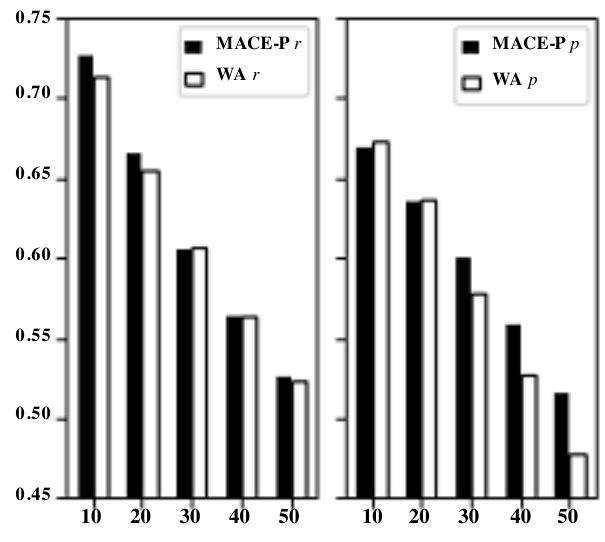}
\caption{Pearson (left) and Spearman correlations of various cut-off percentiles for the BERT-FT\textsubscript{TOPIC} models, trained on data containing \textit{WA} and \textit{MACE-P} quality scores.}
\label{fig:cut_off_correlations}
\end{center}
\end{figure}

An important property of a good model is that its performance increases when considering arguments that are on the extremes of the argument quality scale. To evaluate this, we define a \textit{cut-off} percentile \textit{d} as a view of the test data that considers only the top d and bottom d percent of the data. For each cut-off we calculate Pearson and Spearman correlations w.r.t the predictions of the BERT-FT\textsubscript{TOPIC} model. Figure \ref{fig:cut_off_correlations} presents the correlations for cut-off percentiles ranging from $10\%$ to $50\%$ ($50\%$ is equivalent to taking the entire data), for the models trained on \textit{MACE-P} and \textit{WA} quality scores. A clear trend emerges in which the correlations increase as arguments are taken from a smaller percentile, i.e. from further extremes of the quality scale, reaching up to $.71$-$.73$ and $.67$ for Pearson and Spearman correlations, respectively, when considering only the bottom and top $10\%$ of the test set for evaluation.

\subsection{Experiments on SwanRank and UKPRank}

In this section we demonstrate the strength of the two best methods -- BERT-FT and BERT-FT\textsubscript{TOPIC} -- by applying them on two related datasets, \textit{SwanRank} and \textit{UKPRank}. We chose \textit{SwanRank} as it also contains direct point-wise quality labels. \textit{UKPRank} was chosen as an established dataset which also contains point-wise quality scores. However, it should be noted that as opposed to the \textit{\ourds{}} and \textit{SwanRank} datasets, the point-wise labels in \textit{UKPRank} were induced from a pair-wise labeling task, and not obtained directly, as mentioned previously.

\textbf{SwanRank dataset}. In \citet{swanson2015argument} the \textit{SwanRank} dataset was evaluated both in in-domain and cross-domain scenarios. In this work, we focus on the latter, more challenging scenario. Their model is a SVR with RBF kernel, with features coming from one of many feature sets they experiment with. They randomly split each topic to train ($75\%$) and test sets, and run their model, with each possible feature set, on each train-test cross-domain pair. That is, the topics of the train and the test sets are distinct. We adapt to their approach as follows: 

\begin{itemize}
    \item For each topic, we create our own random split to train and test, as the exact split to train and test for each topic was not available to us.
    \item For each topic, we consider its best result achieved in \citet{swanson2015argument} in the cross-domain scenario. For example, for evaluating the \textit{gun control} topic, the best result in \citet{swanson2015argument} is obtained by training on the \textit{gay marriage} topic. By this we obtain 4 train-test pairs: \textit{gay-marriage} (GM, train)--\textit{gun control} (GC, test), \textit{gun-control}--\textit{gay marriage}, \textit{death penalty} (DP)--\textit{evolution} (EV), and \textit{evolution}--\textit{death penalty}. It should be noted, that adapting to this setup puts our work at a disadvantage, because the best results for different pairs in \citet{swanson2015argument} are achieved by using different feature sets, as opposed to a single learning framework. Thus, this setting represents an upper limit rather than an actual obtained result.
    \item For each of the $4$ pairs, we train our BERT-FT and BERT-FT\textsubscript{TOPIC} methods for $2$ epochs on the train topic, and test the model on the test topic. We compute Root Relative Squared Error (rrse), following \citet{swanson2015argument}, and report results for each pair as well as the weighted-average.
\end{itemize}  

\textbf{Results.} Results on the \textit{SwanRank} dataset are presented in Table \ref{table:resultsSwan}. Using the BERT-FT\textsubscript{TOPIC} method yielded an average improvement of $.8$ points compared to the optimal result taken from \citet{swanson2015argument}. Performance has improved in $3/4$ test topics (\textit{gay marriage},  \textit{gun control} and \textit{death penalty}), and decreased in the \textit{evolution} topic. However, in \citet{swanson2015argument} the performance on the \textit{death penalty} and \textit{evolution} topics is low even in the easier in-domain task, presumably indicating they are much more difficult to predict.

\begin{table}[htb]
\begin{center}
\small
\begin{tabular}{ |l|c|c|c|c|c|  }
 \hline
 Train & GC & GM & EV & DP & Avg\\
 Test & GM & GC & DP & EV & \\
 \hline
 Swanson & $0.84$ & $0.81$ & $1$ & $\textbf{0.97}$ & $0.89$\\
 \hline
 BERT-FT & $0.8$ & $0.82$ & $1.1$ & $0.98$ & $0.91$\\
 BERT-FT\textsubscript{TOPIC} & $\textbf{0.65}$ & $\textbf{0.71}$ & $\textbf{0.96}$ & $1.01$ & $\textbf{0.81}$\\
 \hline
 \end{tabular}
 \end{center}
 \caption{Weighted-average RRSE on the $4$ topic pairs of the \textit{SwanRank} dataset (a lower score is better). \textit{Swanson} row: the result by averaging the best train-test pairs cross-domain results published in \citet{swanson2015argument}.}
\label{table:resultsSwan}
\end{table}

\textbf{UKPRank dataset}. We conduct cross-validation where in each fold we trained on $31$ topics and tested on the held-out topic. The model was evaluated after $5$ training epochs. Following \citet{Gurevych18}, we report average Pearson ($r$) and Spearman ($\rho$) correlations, and compare the results of our methods to the Bi-LSTM and GPPL methods published there, as well as to the EviConvNet method, the best result from \citet{gleizeACL2019}, and to the SWE+FFNN method, the best result from \citet{potashRanking2019}.

\textbf{Results.} The results on the \textit{UKPRank} dataset are presented in Table \ref{table:resultsUKP}.
Both of our methods obtain Pearson correlations which are comparable to the SWE+FFNN, EviConvNet, and GPPL methods, and worse Spearman correlations. However, as the point-wise quality scores were obtained via a pair-wise proxy, previous methods assume the existence of a pair-wise labeled dataset for training, and EviConvNet, for example, trains on the pair-wise labels directly. Our method is less suitable for this setting, as it does not depend on any pairs being labeled for training, making this comparison less trivial.

\begin{table}[htb]
\begin{center}
\begin{tabular}{ |l|c|c|  }
 \hline
 & $r$ & $\rho$ \\
 \hline
 Bi-LSTM GloVe & $.32$ & $.37$\\
 Bi-LSTM ling+GloVe & $.37$ & $.43$\\
 \hline
 GPPL ling & $.38$ & $.62$ \\
 GPPL GloVe & $.33$ & $.44$ \\
 GPPL ling+GloVe & $.45$ & $.65$ \\
 GPPL opt. ling+GloVe & $.44$ & $.67$\\
 \hline
 EviConvNet & $.47$ & $.67$ \\
 \hline
 SWE+FFNN & $\textbf{.48}$ & $\textbf{.69}$ \\
 \hline
 BERT-FT & $.45$ & $.63$ \\
 BERT-FT\textsubscript{TOPIC} & $.46$ & $.62$\\
 \hline
 \end{tabular}
 \end{center}
 \caption{Average correlation on the \textit{UKPRank} dataset.
 }
\label{table:resultsUKP}
\end{table}

\section{Learning to Represent Quality}
\label{analysisModels}

\begin{figure}[htb]
\begin{center}
\includegraphics[]{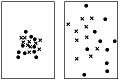}
\caption{A 2D t-SNE projection of the embeddings of the term \texttt{people} 
taken from a sample of 
high (\texttt{x}) and low (\texttt{o}) quality arguments from the dev and test sets of the \textit{\ourds{}} dataset. Left - using BERT's pre-trained model. Right - 
after fine-tuning.}
\label{fig:model_analysis}
\end{center}
\end{figure}

During fine-tuning BERT to a new task, the weights of the pre-trained model are updated, and as a result, the contextual representations of tokens change. In this section, we exemplify how these new embeddings are enriched with contextual quality properties. Using an anecdotal example, we show how quality is encoded in BERT's token embeddings, after being exposed to the \textit{\ourds{}} dataset during training. We do this by comparing the contextual embeddings retrieved from BERT's pre-trained model and from the BERT-FT model, of a common token in the \textit{\ourds{}} dataset, \texttt{people}. For the purpose of this analysis, we use the model fine-tuned for the \textit{\ourds{}} dataset with \textit{WA} scores. First, we split the data to $5$ equally populated bins, according to the \textit{WA} quality score.\footnote{We exclude the training set, to avoid a trivial bias of the BERT-FT model having trained on this set.} We then retrieve the contextual embeddings of \texttt{people} in a sample of $20$ arguments which contain it, $10$ from the lowest and $10$ from the highest quality bins. Embeddings are taken from the last layer of BERT's model before and after fine-tuning, resulting in two sets of $20$ embeddings overall. We cluster each set of embeddings using Hartigan's K-Means \cite{hartigan1975,Slonim18297}, and calculate the adjusted mutual information (AMI) of the clusters with respect to the low and high quality bins. When clustering the embeddings retrieved from BERT's pre-trained model, the AMI is $0.08$, a low result which is expected given that BERT's language model should not have any preference to argument quality. However, when clustering the embeddings extracted from BERT-FT, the AMI reaches $0.31$, indicating that these representations absorb the qualitative nature of the arguments they come from. We use t-SNE \cite{vanDerMaaten2008} to visualize these embeddings in a 2D space (Figure \ref{fig:model_analysis}). Before fine-tuning, the embeddings are clustered together, whereas after fine-tuning, the embeddings are partially separated by the quality of the arguments, further indicating that the quality contributes to the contextual representation of this token.

\section{Conclusions and Future Work}

In the rapidly expanding field of computational argumentation, argument quality is an increasingly prominent issue, having substantial implications. To advance the development of argument quality ranking models, the creation of new datasets is a critical step. In this work, we present a novel argument dataset labeled for point-wise quality, \textit{\ourds{}}, containing $30{,}497$ arguments. To the best of our knowledge, this dataset is the largest to include point-wise quality labels, $5$ times larger than previously released datasets. We follow \citet{Toledo19} by collecting the arguments actively, while employing elaborate annotation control measures. A practical question, 
overlooked 
in previous datasets, is how to induce continuous labels from binary annotations. We address this issue by conducting an extensive comparison of two common 
approaches and analyzing their appropriateness to our dataset. We also 
exploit 
this dataset to the task of argument quality ranking, by presenting a BERT-based neural method which outperforms several baselines. We show this method is capable of achieving promising results on other datasets as well. We believe that the approach to argument collection, the analysis of different labeling scores, as well as the sheer size of the dataset, make it useful for further advancements in this field.

As an attempt to provide insight regarding the characteristics of \textit{\ourds{}}, we conducted an analysis of quality dimensions, showing that the dimensions of \textit{Global Relevance} and \textit{Effectiveness} are the most indicative to overall quality scores. As future work, we would like to further explore this, by investigating how quality dimensions impact overall quality, and whether a prediction model can capture these dimensions effectively.

\section*{Acknowledgements}
We thank Eyal Shnarch, Leshem Choshen, and Elad Venezian for their insightful comments and suggestions.

\fontsize{9.0pt}{10.0pt} \selectfont
\bibliography{aaai_arg_quality}
\bibliographystyle{aaai}

\end{document}